\documentclass{article}
\usepackage{ijcai19}

\usepackage{times}
\usepackage{soul}
\usepackage{url}
\usepackage[hidelinks]{hyperref}
\usepackage[utf8]{inputenc}
\usepackage[small]{caption}
\usepackage{graphicx}
\usepackage{amsmath}
\usepackage{booktabs}
\usepackage{algorithm2e}
\usepackage{algorithmic}

\usepackage{amssymb}

\usepackage{xcolor}

\DeclareMathOperator*{\argmin}{arg\,min}
\usepackage[load-configurations=version-1]{siunitx} % newer version
\usepackage{todonotes}
\newcommand{\overbar}[1]{\mkern 1.5mu\overline{\mkern-1.5mu#1\mkern-1.5mu}\mkern 1.5mu}
\newcommand{\udayannote}[1]{\textcolor{red}{Udayan: #1}}

\newcommand{\eat}[1]{}
 \newcounter{personaln}
 
\title{Automating Predictive Modeling Process using Reinforcement Learning}

\iftrue
\author{
Udayan Khurana
\And
Horst Samulowitz
\affiliations
IBM Research AI\\
\emails
\{ukhurana, samulowitz\}@us.ibm.com
}
\fi

\begin{document}
 
% \author{Udayan Khurana\\
%ukhurana@us.ibm.com\\
%IBM Research AI
%\And
%Horst Samulowitz\\
%samulowitz@us.ibm.com\\
%IBM Research AI
%}

\maketitle

 % The optional argument of \title is used in the header

\begin{abstract}
Building a good predictive model requires an array of activities such as data imputation, feature transformations, estimator selection, hyper-parameter search and ensemble construction. 
Given the large, complex and heterogenous space of options, off-the-shelf optimization methods are infeasible for realistic response times. 
In practice, much of the predictive modeling process is conducted by experienced data scientists, who selectively make use of available tools. Over time, they develop an understanding of the behavior of operators, and perform serial decision making under uncertainty, colloquially referred to as educated guesswork.  
With an unprecedented demand for application of supervised machine learning, there is a call for solutions that automatically search for a good combination of parameters across these tasks to minimize the modeling error. 
We introduce a novel system called {\em APRL} (Autonomous Predictive modeler via Reinforcement Learning), that uses past experience through reinforcement learning to optimize such sequential decision making from within a set of diverse actions 
%(feature construction, model training, hyper-parameter optimization,  ensemble) 
under a time constraint on a previously unseen predictive learning problem. 
APRL actions are taken to optimize the performance of a final ensemble. This is in contrast to other systems, which maximize individual model accuracy first and create ensembles as a disconnected post-processing step. As a result, APRL is able to reduce up to 71\% of classification error on average over a wide variety of problems.
\end{abstract}

\section{Introduction}
%\tagnote{general}
In recent years, applied machine learning has received significant attention. 
There is an increasing availability of digitally recorded and disseminated information that can be used to build predictive models in various domains across the industry, academia and government. 
The building of reliable predictive models is a task primarily conducted by data science teams. 
The primary objective of a data scientist is to try various tools and techniques at her disposal, aggregate the effective operations based on results, fine tune them, and finally test before deploying a generalizable predictive modeling solution. 
%\udayannote{Mention indispensable skills of data scientists}
The downside of a data scientist-centric approach is the high cost and high turn around time. There is a significant supply-demand deficit of data scientists~\cite{dsarticle}. Moreover, not every small organization can afford a highly qualified team of data scientists.% and moreover, there is a lack of availability as well.%~\cite{xxx}.

%There is an increased focus on additional tasks to improve accuracy beyond simply cleaning and training with a suitable algorithm. 
%Data scientists often spend hours to days engineering the feature space, optimizing training hyper-parameters, constructing ensembles and so on. 
%\udayannote{ Elaborate this paragraph. Talk about different domains etc.}

%\tagnote{Automl}
This has lead to an increased demand for automated machine learning (AutoML) techniques that computationally solve some or all of the steps involved in the applied machine learning process. 
%In fact, some progress has been registered in the AutoML space, which can be categorized into two areas.
%\udayannote{Non continuous variables}  
AutoML as an optimization problem is challenging. It involves a vast and diverse space of options, making it computationally prohibitive to run optimization solvers for obtaining solutions in realistic time frame using affordable resources. For instance, feature transformations for a single feature, using a modest number of unary functions, say 30, with a combination depth of 3, itself leads to $3^{30}$ configurations to chose from; Sklearn's Random Forest Classifier provides 17 hyper-parameters with varying range of possible values -- some continuous, some discrete; selecting the optimal ensemble base models from a set of many constructed models, is yet another combinatorial problem, and so on. 
%Therefore we expect to see different approaches on this topic, using creative aspects. 
One alternative direction is the use of historical information obtained from model building exercises. Different works recently have been able to learn useful clues from historical information to short-circuit the search on the vast space. For instance, in Bayesian Learning, priors on various options are learned, in reinforcement learning-based techniques, pruning or search strategies are tuned using prior runs, or associating feature distribution characteristics with suitable actions, etc. 
%Firstly, solutions that address individual components such as feature engineering alone without the need for jointly determining the optimal choices for other tasks such as estimator selection.

%\tagnote{automl-specific}
Another direction for AutoML is based is the use of approximations and heuristics for specific tasks such as feature engineering, without considering other aspects such as estimator selection. 
In this context, it should be noted that the choice of learning algorithm often greatly influences the choice of effective feature engineering transformations, and vice versa. 
While the collective use of such individual optimizers may not aim towards a global optimal solution for the ML choice optimization, they are used effectively by practitioners. 
Since these techniques are usually based on task specific heuristics or approximations, it is difficult to integrate them in a unified framework. 
The proposed approach in this paper dwells on the idea of using both -- historical information and task-specific heuristics algorithms. 

\eat{
The other category of AutoML techniques are those that aim to perform a joint optimization of several choices through Bayesian Optimization or Genetic algorithms. The advantage of such approaches is that they aim at finding the joint best solution in theory. However, they are quite time consuming. Also, they may not easily utilize the custom optimizers designed for specific subtasks.% through heuristics. 
}
%\tagnote{kaggle}
%Consider the recent Kaggle competition on home credit loan approval~\footnote{\url{https://www.kaggle.com/c/home-credit-default-risk}} in which more than $7,000$ teams participated and over a duration of three months more than $300,000$ predictions were submitted to the corresponding leaderboard. More interestingly, the difference between the performance of the rank 1st and rank 500th submissions is $0.01$ unit of AUC. By the account of successful Kagglers, final ensembles enable crucial boost in performance. 

%\tagnote{Our  approach Details}
% \udayannote{entry of ensembles seems sudden}

%ML exercise to find the most suitable ensemble for the given problem. 
%While ensembles have been known to be inevitably useful in practice, the idea of a final ensemble is general enough to include the possibility of a single model as the optimal solution. 
In the process of trying different options such as a particular feature transformation, using a particular estimator or certain choice of hyper-parameters, produce several transformed versions of the data or a slightly different model. Amongst several such trials on the way, only one or a few models constructed are deemed winners and adopted. However, upon carefully selecting a subset of these subpar models, simple but effective ensembles can be created that outperform the impact of the best one alone, owing to their mutual diversity besides the individual strength.  
APRL goes a step further and not only makes ensembles from the subpar models obtained in the process, but also explicitly creates models that maximize the performance of the final ensembles.
%Additionally, it is guided with the objective of  finding the best encapsulating ensemble to the given ML problem. This marks a significant difference than creating ensembles as a post-processing step based on finding good models.
% through a careful formula involving individual accuracy and mutual diversity amongst models. 
We use reinforcement learning (RL) to device an exploration strategy that guides the process to take actions that minimize the {\em ensemble generalization error (EGE)} of a proposed final ensemble. Minimizing the EGE directly results in finding a combination of high accuracy and highly diverse models, which is the recipe for a successful ensemble. To this end, we present a fast, greedy algorithm to select a subset of models from a large subset, that minimize EGE. 
In this paper, we describe the core ideas behind APRL, which includes modeling the action space, exploration process, rewards through ensemble generalization and policy learning through reinforcement learning.

\section{Related Work}
%\tagnote{AutoSklearn}
In existing work on AutoML, Auto-sklearn~\cite{automl} and Auto-WEKA~\cite{autoweka1,autoweka2} use sequential parameter optimization based on Bayesian Optimization to determine effective predictive modeling pipelines by combining data pre-processors, transformers and estimators. Both variants are based on the general purpose algorithm configuration framework SMAC~\cite{smac} to find optimal machine learning pipelines. In order to apply SMAC, the problem of determining the appropriate ML approach is cast into a configuration problem where the selection of the algorithm itself is modeled as a configuration.
Auto-sklearn also supports warm-starting the configuration search by trying to generalizes configuration settings across data sets based on historic performance information.
Auto-sklearn is of particular interest in the context of this paper as it constructs an ensemble of classifiers instead of a single classifier. It uses ensemble selection from~\cite{ensembleselection} which is a greedy algorithm that starting from an empty set of models incrementally adds a model to the working set in each step if it results in maximizing the improvement of predictive performance of the ensemble. 
%\udayannote{ensemble in postprocessing?}

%\tagnote{Genetic algos}
Another approach for automated ML is based on genetic algorithms~\cite{olson2016tpot}. While it does not create ensembles, it could compose them based on derived models as the authors point out. More interestingly, the approach presented in~\cite{geneticensembles} uses multi-objective genetic programming to evolve a set of accurate and diverse models via biasing the fitness function accordingly.

%\tagnote{FE general}
There is a diverse set of approaches towards automated feature engineering which are summarized below.
FICUS~\cite{MarkovitchRosenstein02} performs a beam search over the space of possible features, constructing new features by applying constructor functions. Its is guided by heuristic measures based on information gain in a decision tree.
%, and other surrogate measures of performance. 
Data Science Machine by~\cite{kanter:dsm} applies all transformations on all features at once (but no sequence of transformations), then performing feature selection and model hyper-parameter optimization over the augmented dataset. FEADIS~\cite{DorReich12} works through a combination of random feature generation and feature selection. 
%It adds constructed features greedily, and as such requires many expensive performance evaluations. 
ExploreKit~\cite{DBLP:conf/icdm/KatzSS16} expands the feature space explicitly, one feature at a time. It employs learning to rank the newly constructed features and evaluating the most promising ones. 
LFE~\cite{lfe},~\cite{aaai18demo} directly predicts the most useful transformation per feature based on learning effectiveness of transforms on sketched representations of historical data through a perceptron. Cognito~\cite{khurana2016cognito},~\cite{afe} presents a rule based hierarchical exploration of transforms. ~\cite{sondhi:2009} and ~\cite{fechapter} summarize some of the work in feature engineering over the years.
%It considers features independent of each other; it is demonstrated to work only for classification so far, and does not allow for composition of transformations. 
%A combination of the learning-based and heuristic tree-based exploration approaches has also been suggested~\cite{autfenipsw}. 
%A detailed explanation including relationships between these approaches can be found in~\cite{fechapter}.
%While this approach is more scalable than the expand-select type, it still is limited due to the explicit expansion of the feature space, and hence time-consuming. For instance, their reported results were obtained after running FE for days on moderately sized datasets. Due to the complex nature of this method, it does not consider compositions of transformations. 
%Hyper-parameter optimization has also been employed to some limited settings of feature engineering, such as ~\cite{automl}.

%In recent years, different approaches have been proposed for performing automated feature engineering. One particular approach by~\cite{khurana2016cognito} 
%\tagnote{FE Cognito}
The most relevant feature engineering approach to the work presented in this paper is~\cite{khuranaaaai18}, which
is based on trial or exploration of different transformation functions and finding sequences with higher returns based on initial feedback. Their trials are organized in a hierarchical, directed acyclic graph and the goal is the minimization of model error. Their exploration policy is trained through reinforcement learning on historical ML problems. In this paper, we also employ a hierarchical structure for exploration, with a much diverse space of actions including ensembles, hyper-parameters, estimator selection. Along with that, our reward mechanism reflects the need to optimize for ensemble goals of both, high accuracy and diversity instead of model accuracy alone, and use multiple estimators instead of just single one. 
 
%In recent years, we have witnessed various efforts to perform automated feature engineering/construction.~\cite{khurana2016cognito} introduce the notion of a tree-like exploration of transform space; they present a few simple handcrafted heuristics traversal strategies such as breadth-first and depth-first search that do not capture several factors such as adapting to budget constraints. Those concepts are generalized by ~\cite{khuranaaaai18}, using Q-learning to find an efficient strategy to explore a large number of feature engineering options through a hierarchical structure. In this work, we extend the feature engineering framework by ~\cite{khuranaaaai18}, to optimize for ensembles instead of only feature engineering performance.

%We refer to this FE approach as {\em evolution-centric}. 

%\subsection{Ensembles}
%\tagnote{Ensembles}
Model ensembles are used extensively in machine learning to aggregate the output of several weak predictors into a single strong predictor~\cite{Dietterich2000}. 
%Ensembles can be constructed in many different ways. The basic ask, however, is that the base predictors perform above a threshold (above random chance, such as $p > 0.5$ for binary classification), and be diverse enough. 
Different methods to ensemble ensure diversity through different means. For example, Random Forests by ~\cite{breiman1999random} use systematic randomization to create data subsets and perturbation in branch splitting to reduce variance in a large number of decision trees. The aggregation is a simple unweighted average.% the predictions of all base models.
%\udayannote{ More on Ensemble pruning and Ensemble learning.}

%\tagnote{misc}
%\udayannote{Try to move away from FE centric discussion}
In the given context, we area dealing with specific ensemble needs. Due to the associated cost of exploration involving model building and validation, there is a moderate number of base models (typically 10-100), all of which are fairly correlated to each other, with small differences. The mutual correlation is because all of them contain an overlapping set of (original) features, apart from different transformed features, which often causes mild to moderate deviation in models' behavior, or none. Hence, a simple averaging of all base models seems ineffective, and a rather careful selection of base models is required. We also observe that performance oriented exploration can often overfit the training data, but presents a blessing in disguise for ensembles, when handled appropriately~\cite{Sollich1996}. We acknowledge that other known mechanisms to improve model diversity can be used along with our proposed methods, such as the use of different hyper-parameter configurations, random subset selection, amongst others. In fact, these mechanisms are complimentary to the core idea suggested in this paper. 
%However, in order to evaluate the merit of the core technique proposed here, we stick to the choice of a single learning algorithm with a fixed set of hyper-parameters and do not include data sub-selection explicitly (it may happen within a learning algorithm). 
To further emphasize the benefits of the core approach, we present promising results with simple averaging instead of employing more sophisticated approaches such as stacking \cite{Gunes2017} or boosting \cite{weaklearners}, that are obvious additions to our ongoing work.
%Stacked Ensemble Models for Improved Prediction Accuracy~\cite{Gunes2017}.
%\begin{pnote}
%Why does it have to be a moderate number of base models? One could run exploration for a long time.
%Also, why do the models need to have small difference. I see why its likely that they do but its not necessary. 
%In addition one can have a different model per data set and therefor diverse results (and maybe that is why random works quite well too?) 
%What about stacking?
%\end{pnote}

%According to Condorcet's jury theorem:
%\textit{
%If p is greater than 1/2 (each voter is more likely to vote correctly), then adding more voters increases the probability that the majority decision is correct. In the limit, the probability that the majority votes correctly approaches 1 as the number of voters increases.}
 %\begin{pnote}
 %      Write a proper paragraph for some of the following references.
  %  \end{pnote}

%Explicit emphasis on diversity for ensembles is abundant.
%\tagnote{Diversity for ensembles}
There exists a body of work on generating diversity for ensembles.~\cite{Melville2003}, for instance, generate artificial examples to construct diverse predictors whose ensemble provides good gains. ~\cite{Cunningham2000} and ~\cite{Zenobi2001} introduce diversity based on feature selection through hill climbing algorithms. A survey of diversity promoting methods is presented by~\cite{Brown2005}. 
Reinforcement learning has been used as a metalearning too in the domain of deep neural network tuning recently~\cite{finn2017model}. However, it should be noted that there are significant differences in the structure of deep learning tuning problem compared to that of regular predictive modeling that involves different learning algorithms, explicit and more diverse feature engineering, etc. We similar approach to AutoML for maximizing ensembles through feature engineering was using reinforcement learning~\cite{icmlw} but not performing estimator selection and hyper-parameter optimization.

%Popular Ensemble Methods: An Empirical Study~\cite{Opitz1999}
%Error correlation and error reduction in ensemble classifiers~\cite{Classifiers1996}

%Ensemble Methods in Machine Learning~\cite{Dietterich2000}
%Neural network ensembles, cross validation, and active learning~\cite{Krogh1995}

%\begin{equation}
%    \label{simple_equation}
%    \alpha = \sqrt{ \beta }
%\end{equation}

\section{APRL Overview}
%\udayannote{Either in intro or here, there needs to be a discussion on the process of data science - exploration etc., and how things depend on the given problem are unpredictable. Therefore, handiling the process require adaptive/reactive traits. That is what makes a data scientist so essential in the process. The problem can be thought of as what to try next -- what is most likely to benefit me in the longer run. This approach hasn't been looked at. We solve this problem from RL angle.}
%\udayannote{Explain the action, state space in detail here}

%\udayannote{This or an immediately following section should contain the details on RL}

%In this section, we present the general overview of our approach. In the further two sections, we provide details on the ensemble generalization error, and reinforcement learning based exploration strategy, respectively. 

\begin{figure}[h]
    \centering
    \includegraphics[width=0.45\textwidth]{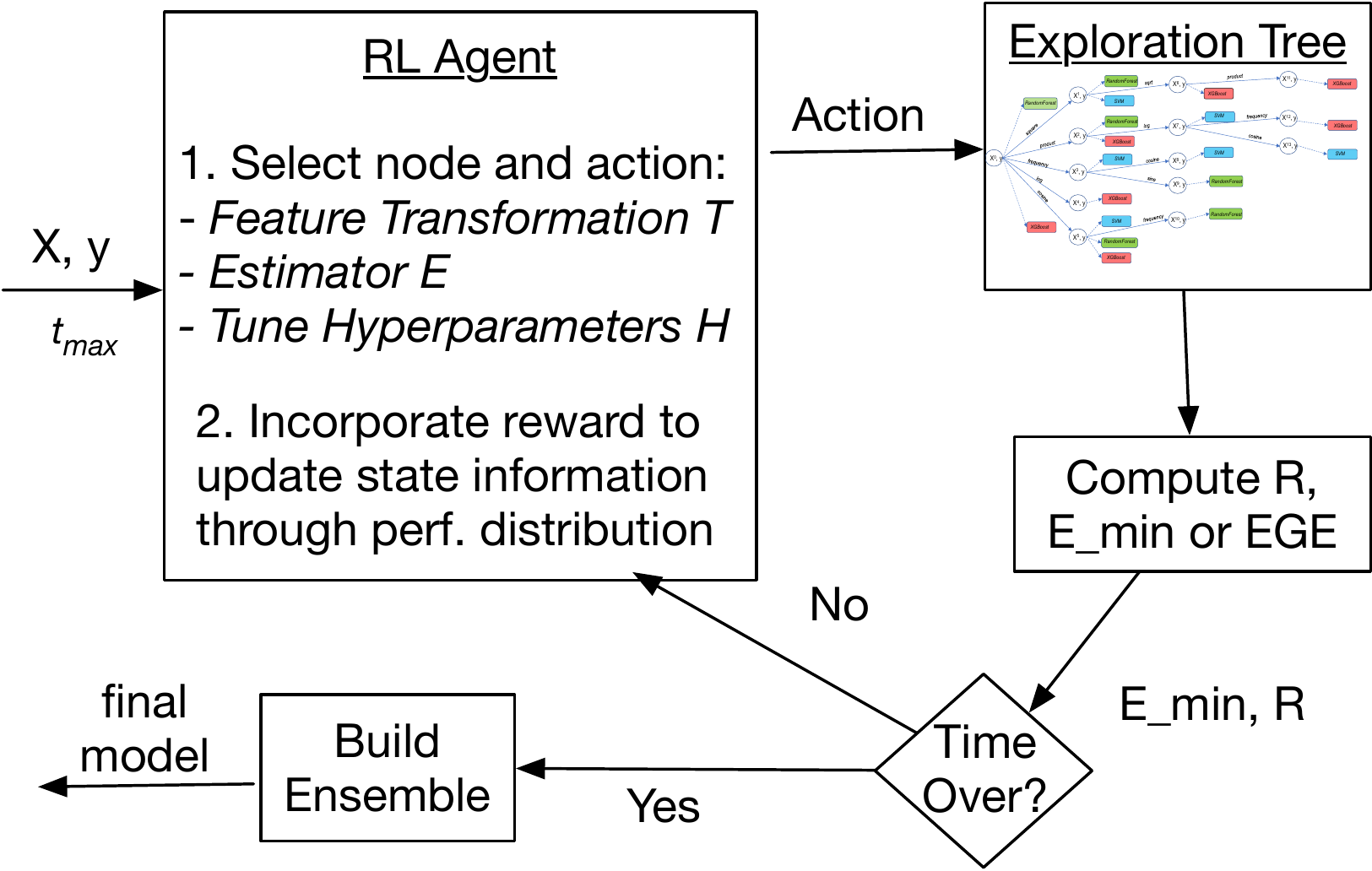}
    \caption{APRL approach Overview. Given a supervised ML problem ($X, y$) and a time constrain $t_{max}$, the system generates an ensemble after iterative exploration through multiple transforms, estimators and hyper-parameters.}
    \label{overview}
\end{figure}

\begin{figure*}[t]
    \centering
    \includegraphics[width=0.85\textwidth]{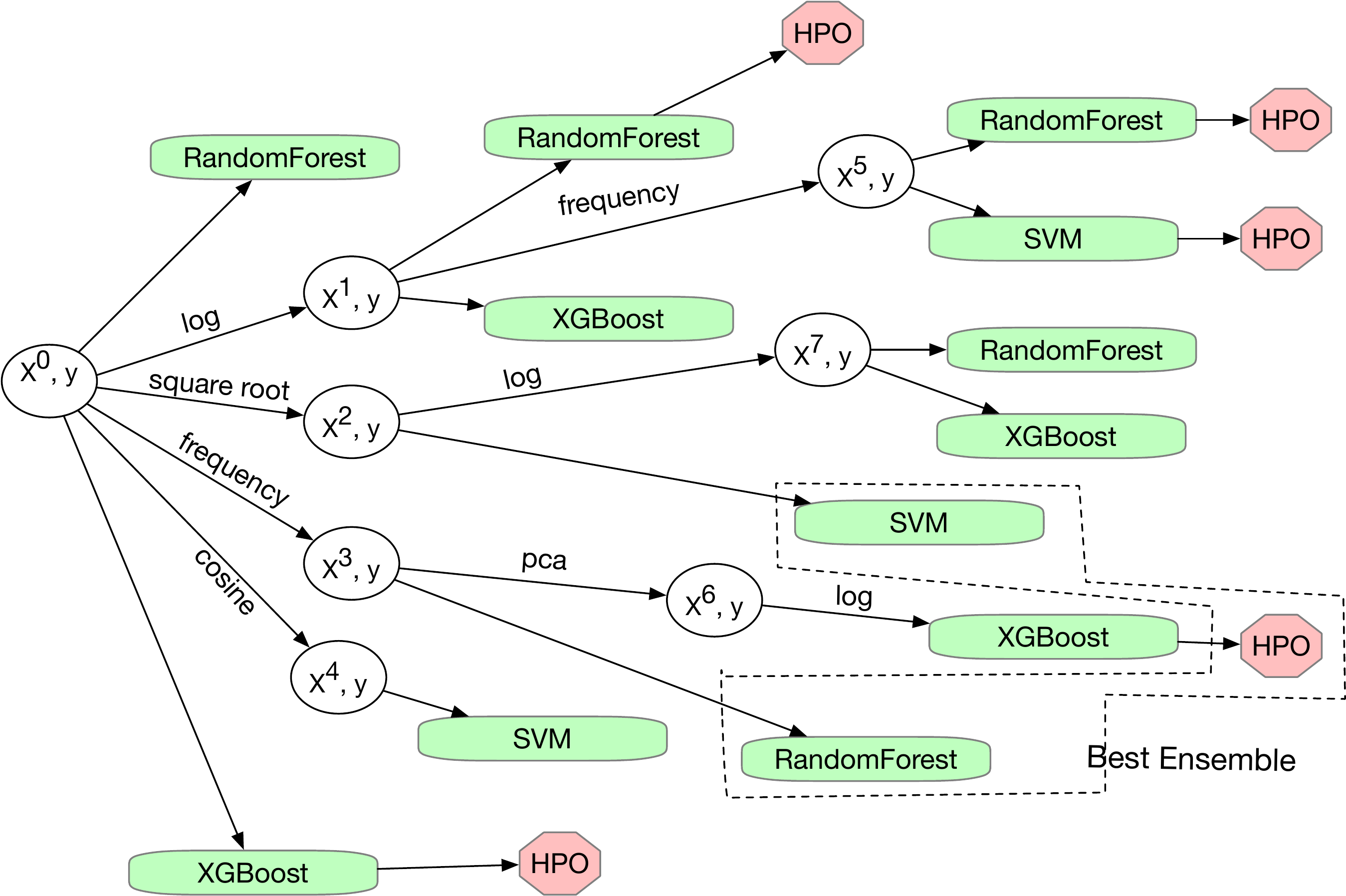}
    \caption{Illustration of an Exploration Tree for a given problem of $X^0, y$. Circular nodes depict data versions owing to transformations shown on the edges. Rectangular nodes correspond to selected estimators. Hexagonal red blocks represent hyper-parameter optimization, and dotted cluster is an ensemble.}
    \label{new_tree}
\end{figure*}

Figure~\ref{overview} illustrates APRL's approach. The cornerstone of APRL is an agent that iteratively decides the action to perform, based on two factors -- the performance of various actions until that moment and time remaining until when the problem must be solved. 
% \udayannote{Mentioning policy learning here abruptly might be confusing wrt to what follows}
%In the rest of this section, we describe the structure of the solution, followed by details in the next two sections.
APRL is presented with a classification or regression problem in terms of feature vectors ($X$) and a target vector ($y$), and the time constraint ($t_{max}$) in which to build a model. 
%The distribution of $y$ automatically determines whether it is a classification or a regression problem. 
In each iteration, the agent has to decide to perform an action from a choice of many available ones. It can apply one of the following, to either the given data (X, y) or a version of it that has been derived in the course of this exploration: (a) feature transformations; (b) estimator selection and model building; (c) hyper-parameter optimization for a model and transformation.
The process of taking such actions is called {\em data science exploration} or simply {\em exploration}, which is illustrated in Figure~\ref{new_tree}. Due to the hierarchical nature of the abstract representation, it is called an {\em exploration tree}.

%At those particular nodes of this tree which contain a model, an

% the following: dictating the choice of feature transformation to be applied and the estimator(s) to be used in each iteration. Additionally, it is decided whether to apply a step for HPO (hyper-parameter optimization)
% This choice is based upon a model which incorporates the current state of the run and the feedback obtained in form of ensemble generalization error (EGE), and the remaining number of estimated iterations, based on remaining time.

%\udayannote{Reverse the order. Describe the technique first then explain why. Possibly isolate the why somewhere else}
Applying a feature transformation to a feature-set means generating new features by applying the corresponding function on all possible input features (or feature combinations for non-unary functions), and appending them to the original feature set. For example, applying a $logarithm$ transform,  $\log(X)$ means creating new features by $\log(x_i) \forall x_i \in X$, such that feature $x_i$ contains only positive values. Similarly, {\em frequency} transform function, that creates a new feature by counting the number of occurrences of a value is done only for discreet values transforms, and so on.  
As described in the related work section, in general, feature transformation can be performed in several ways. It may be done iteratively by adding one new feature at a time using one transform, or in contrast, by adding them in bulk by applying multiple transforms on all given features, usually followed by an intense feature selection process. We adopt a middle of the road approach where, in one iteration, we apply one transform to all valid input feature candidates. 
In this paper, we present results using a set of 9 different transformation functions: {\em freq, pca, round, minmaxscaler, tanh, groupby+stddev, cbrt, sigmoid, stdscaler}, and a special {\em feature selection} transformation that can remove unnecessary features from the input. The selection is based on historical performance,. The approach is independent of the the actual transformations used and additional or fewer transformations may be specified, including domain specific ones.

The second category of actions is the choice of a particular estimator, such as {\em Random Forest}, {\em Gradient Boosted Trees}, {\em K-nearest neighbors}, etc., to be chosen on a transformed (or original) data with its default hyper-parameters, which has not been done so far. The third category of actions is that of a hyper-parameter optimization routine to be run on for data node in the tree. We use a hyper-parameter optimization routine based on black-box optimization using radial basis functions~\cite{costa2018rbfopt}. The action specifies a data node, the estimator, hyper-parameter optimization, and the allocated time for it. Additionally, the result of each hyper-parameter optimization step is treated as a starting value set for the next action belonging to the same estimator type. 

At each iteration, the agent enumerates all actions possible from the current state of the tree (without repetition) and ranked as per the expected long term benefit. The action with the highest expected benefit is chosen and applied. From one state of the tree to the next, the list of possible actions changes. Additionally, the perceived benefit of the actions change as well. The task of taking the best action, given the state of the exploration is performed by the agent based upon a policy, $\Pi$ that it learns historically on other learning problems using reinforcement learning. Further details on the agent's policy learning, definition of states, and such provided in Section~\ref{sec:rl}.

At each node with an estimator, plain or with hyper-parameter optimization, an prediction of the labels is performed through 5-fold cross validation. The predictions thus generated are required to compute the ensemble generalization error (EGE) for a potential ensemble that may contain that node. It is further used to compute $E_{min}$, a quantity that reflects the minimum value of EGE for an ensemble from any subset of the nodes of a given tree. This quantity is used as a feedback reward for the agent to learn its policy. Further details are presented in Section~\ref{sec:ensemble}.

%The RL agent is trained on a historical set of data, and the exploration strategy has been tuned to obtain the maximum amount of diversity as well as accuracy from the models generated in a number of iterations ($N$). In a nutshell, for a given data, the agent first explores different transformations and estimators and their combinations, and then identifies further trials to be conducted within the given budget of iterations, which would minimize the potential EGE, as if an ensemble were to be formed.
%\udayannote{Provide examples of what RL can help with and regular optimization doesn't}

%The ensemble is formed by simple averaging. In the case of regression, the predictions are averaged, whereas in case of classification, the probabilities for respective class predictions are averaged amongst different base models, followed by assigning classes upon thresholding. The ensemble is created by picking a subset of the available models that minimize the EGE criterion. More complex ensembles may be used in place of averaging. However, average based ensembles enable us to compute the EGE all along the process quite cheaply, allowing us to perform more in terms of exploration of feature engineering and estimator training. 

\eat{
\begin{figure}
    \centering
    \includegraphics[width=0.50\textwidth, height=5cm]{tree.pdf}
    \caption{Illustration of an Exploration Tree for a given problem of $X^0, y$. \udayannote{ Redo the diagram with ensembles and HPO}}
    \label{tree}
\end{figure}
}

% Also, our proposed general approach can work with other specific alternatives for the RL model, feature transformation function set, validation method (train/test split, for example), and another suitable ensemble performance measure in place of EGE. 

 \eat{                  
\section{Metalearning by Agent}
In this section, we present the details of the exploration process and obtaining the exploration policy through reinforcement learning. 
Our approach resembles the general idea for feature engineering presented by~\cite{khuranaaaai18} \udayannote{Remove} with significant differences, however. We fulfill the objective of minimizing the ensemble error instead of an individual model's error. We consider the space of  estimator algorithms as well and not only feature transforms. Finally, our exploration requires a simpler tree structure instead of a directed acyclic graph. 
}

\section{Learning Exploration Policy for Agent}
\label{sec:rl}

We model the exploration process as a Markov Decision Process (MDP), where a state of the system contains all the required information to take the next action at any point. 
The {\em state} at any iteration is described by a combination of the exploration tree state (entire tree up to that point with all details such as gain in model accuracy due to transforms, estimators, EGE, etc.) and the fraction of remaining time $(t_{max}-t)/t_{max}$ where $t=0$ at the beginning.
%at step $i$ is a combination of two components: (a) transformation tree snapshot with $i$ nodes; (b) the remaining budget at step $i$, i.e., $b_{ratio} = \frac{i}{B_{max}}$. 
%Note that a state of $\mathcal T$ implies the knowledge of impact of various transformations, which play a role in the definition of the state. 
Let the entire set of states be $S$. An action for this MDP is the application of a particular transformer, estimator or HPO on a node of the tree at that state.
%On the other hand, an {\em action} at step $i$ is a pair of existing tree node and transformation, i.e., $<n,t>$ where $n$ is a $node(G_t)$, $t \in T$ and $\nexists n' \in G_i$ such that $edge(n, n') = t$; it signifies the application of the one transform (which hasn't already been applied) to one of the exiting nodes in the graph. 
Let the entire set of actions be $C$. A policy, $\Pi : S \rightarrow C$, guides which action should be taken for a given state. We first learn $\Pi$ iteratively through RL on a set of many prediction problems, and apply it to use for an unseen data. %here is to learn the optimal policy (exploration strategy) by learning the action-value function, which we elaborate later in the section.

The ``remaining time budget fraction'' is a part of the state definition because it is used to determine whether to explore more (i.e., experiment with different transformations and estimators, usually at the beginning of a run) or exploit more (i.e., focus more towards steps that will likely give a desired result based on the exploration so far). 
%Such formulation uniquely identifies each state of the MDP, including the context of ``remaining budget'', which  
%helps the policy implicitly play an adaptive {\em explore-exploit} tradeoff. 
It decides whether to focus on exploiting gains (depth) or exploring (breadth) or a compromise, in different regions of the graph, at different steps. 
%The same action may be assigned different relative values by a policy in the context of different remaining budget values. 
Overall, the policy selects the action with the highest expected long-term reward contribution; however, upon evaluating a new node's actual immediate contribution, the expectations are often revised and explore/exploit gears are (implicitly) calibrated through the policy. 
For example, upon finding an exceptionally high improvement at a node during early stages, the (breadth) exploration can be temporarily localized under that node instead of the same level as it.
Overall, value estimation is a function based on multiple attributes of the MDP state such as current remaining budget, tree structure and relative performance at various nodes, etc. 
Note that this particular {\em explore-exploit} tradeoff is implicitly learned as a part of the policy. 
Note that this runtime explore-exploit tradeoff is different from the explore-exploit tradeoff seen during RL training in context of selecting actions to balance reward and not getting stuck in a local optimum, which is controlled explicitly. We employed an $\epsilon$--greedy methodology, where an action is chosen at {\em random} with probability $\epsilon$ (random exploration), and from the {\em current policy} with probability $1-\epsilon$ (policy exploitation). The trade-off is exercised randomly and is independent of the state of MDP. The value of $\epsilon$ is constant and is chosen based on experimentation.

At step $i$, the occurrence of an action results in a estimator node, $n_i$, and the best ensemble error $E_{min}(n_i)$ for a subset of of nodes $\{0, 1 \dots n_i\}$ is obtained using Algorithm~\ref{alg:greedy1}. 
To each step, we attribute an immediate scalar reward (with a slight abuse of notation):
\[r_i ={{ {E_{min}(nodes(\mathcal T_{i-1}))} -  {E_{min}(nodes(\mathcal T_{i}))}} \over E_{min}(\{X_0\}) }\]
with $r_0=0$, by definition. Also, $E_{min}(\{X_0\})$ is the 5-fold CV performance using any particular estimator. 
%The immediate reward signifies the contribution of the corresponding action if the final solution 
%The total reward at the end of a run is defined as  $r_{final} = \max_{n \in \theta(G(i_{final}))}{A(n)} - A(n_0)$.
 The cumulative reward over time from state $s_i$ onwards is defined as $R(s_i) = \sum_{j=0}^{N}{\gamma^i. r_{i+j}}$, 
where $\gamma \in [0,1)$ is a discount factor, which prioritizes earlier rewards over the later ones. Here, we find the optimal policy $\Pi^*$ which maximizes the cumulative reward. % for a given dataset.
%To each step, we attribute a scalar reward which is an estimate of how much we attribute the result of this action towards the success of the entire exploration process. 
%The utility of a reward is to optimize the sequence of actions that maximize the overall reward in the process. 
%The notion of a step's contribution to the overall task is not known at the moment, and instead we work with a notion of an immediate reward. 
%An intuitive measure of the reward is simply the improvement in accuracy obtained from the addition of this node compared to the accuracy of it parent node,  $r_t = A(N_t) - A(N_t^p)$, where $N_t^p$ is the parent of $N_t$. Another measure of reward is the difference between the accuracy of the new node and the best node so far,  $r_t = A(N_t) - \max_{i = 0 }^{t-1}{A(N_i)}$. We omit a more detailed discussion of the reward formula, and use a mean of the two examples as a reasonable working example for the rest of this section.
The general result of Q-learning~\cite{watkins1992q} to learn the action-value Q-function, for each state, $s \in S$ and action, $c \in C$, with respect to policy $\Pi$ is: % defined as: 
\[Q(s, c) = r(s, c) + \gamma R^\Pi {(\delta(s,c))} \]
 where $\delta : S \times C \rightarrow S$ is a hypothetical transition function, and $R^\Pi(s)$ is the cumulative reward after state $s$. The optimal policy is found as: 
\[ \Pi^*(s) = \operatorname*{arg\,max}_{c}[Q(s,c)] \] 
Due to the extremely large state space, $|S|$, it becomes infeasible to learn the Q-function directly. Instead, we utilize a linear approximation for the Q-function, %similar to Iradova et al.~\cite{irodova2005reinforcement} as follows:
$Q(s,c) =  w . f(s)$, 
where $w$ is a weight vector and $f(s)$ is a vector of the state characteristics such as: (a) average model performance for a circular node (average of all estimators through it); (b) best performance for a circular node ; (c) Average performance of an estimator type; (d) average gain in accuracy for a transform; (e) best gain in accuracy for a transform; (f) Remaining time; (g) Total allocated time (h) Count of numerical features; (i) Count of categorical features; (j) Whether it contain date/time; (k) Remaining time budget fraction times value of (a); (l) Remaining time budget fraction times value of (b), and so on.  

Hence, we approximate the Q-function values with linear combinations of characteristics of a state of the MDP. 
The update rule for $w$ is as follows, where $g'$ is the state of the tree at step $j+1$, and $\alpha$ is the value for learning rate:
\begin{equation} \label{eq:1}
w^{j} \gets w^{j}+ \alpha . (r_j + \gamma . \max_{c \in C} Q(g',c') - Q(g,c)). f^{j}(s) 
\end{equation}

 %The proof follows from~\cite{irodova2005reinforcement}.

%A variation of the linear approximation where the coefficient vector $w$ is independent of the action $c$, is as follows:
%\begin{equation} \label{eq:approx2}
%Q(s,c) =  w . f(s)
%\end{equation}

%This method reduces the space of coefficients to be learnt by a factor of $c$, and makes it faster to learn the weights. It is important to note that the Q-function is still not independent of the action $c$, as one of the factors in $f(s)$ or $f(g,n,t,b)$ is actually the average immediate reward for the transform for the present dataset. Hence, Equation~\ref{eq:approx2} based approximation still distinguishes between various actions ($t$) based on their performance in the transformation graph exploration so far; however, it does not learn a bias for different transformations in general and based on the feature types (factor \#9). We refer to this type of strategy as $RL_2$. In our experiments RL2 efficiency is somewhat inferior to the strategy to the strategy learned with Equation~\ref{eq:approx}, which we refer to as $RL_1$.

\section{Ensemble Generalization Error}
\label{sec:ensemble}
We now describe the criterion to compute the ensemble generalization error (EGE) for a given set of base models and an algorithm to select a subset from a given set of models thats maximizes the EGE. While it is used to create a final ensemble, more importantly it is also used to compute a reward for the RL training and runtime stages as described in the previous section.
Krogh et al.~\cite{Krogh1995} provide a useful expression for computing the ensemble generalization error. For a number of base models and the output of model $\alpha$ on input $x$ be $V^\alpha(x)$. Let a weighted average ensemble output on $x$ be, 

\[ \overbar{V}(x) = \sum_{\alpha}{w_{\alpha}{V^{\alpha}(x)}} \] 

The ambiguity on input $x$ of a single member of the ensemble is defined as $a^\alpha = (V^{\alpha}(x) - \overbar{V}(x))^2$. The overall ambiguity of the ensemble on input $x$ is:

\[ \overbar{a}(x) = \sum_{\alpha}{w_{\alpha}a^\alpha(x)}  =  \sum_{\alpha}{w_{\alpha}(V^{\alpha}(x) - \overbar{V}(x))^2} \]

This is the variance of the weighted ensemble around the weighted mean. Let $y(x)$ be the true outcome value for input $x$.  The squared errors for the model $\alpha$ and the ensemble respectively are: $ \epsilon^{\alpha}(x) = (f(x)-V^{\alpha}(x))^2 $, and
$ e(x) = (f(x)-\overbar{V}(x))^2 $. Now, let the weighted average error of the models be $\overbar{\epsilon}(x) = \sum_{\alpha}{w_{\alpha}\epsilon^{\alpha}(x)}$. By rearrangement, we obtain, $e(x) =\overbar{\epsilon}(x) - \overbar{a}(x)$. Averaging the above over several inputs:

\vspace{-5pt}
\begin{equation}
    \label{eq:6}
    E = \overbar{E} - \overbar{A}
\end{equation}

which states that the generalization error of the ensemble equals the weighted average of the generalization errors of the individual models ($\overbar{E} = \sum_{\alpha}{w_{\alpha}E^{\alpha}}$) minus the weighted average of the ambiguities of the individual models ($\overbar{A} =  \sum_{\alpha} w_{\alpha} A^{\alpha}$). $E^\alpha$ and $A^\alpha$ are model $\alpha$'s average error and ambiguity, respectively.
%\subsection{0/1 loss based ambiguity}
%
%\[
%    a^{\alpha}(x)= 
%\begin{cases}
%    0 & \text{if } classV^{\alpha}(x)=class\overbar{V}(x)\\
%    0,              & \text{otherwise}
%\end{cases}
%\]

%\subsection{Diversity based on conditional-entropy}

%\subsection{Algorithm to Select Ensemble Constituents from a Pool of Correlated Models}
%\label{sec:algo}
The significance of the result by~\cite{Krogh1995} in Equation~\ref{eq:6}, is that it can help evaluate the relative performance of different model sets through individual model performance and a measure of average ambiguity. However, selecting the best subset of models that minimizes this error is a combinatorially hard problem. Knapsack-type approximation algorithms do not  work in this case because we need to minimize the difference of two unrelated quantities. We also explored set coverage style of solutions but found them to be inadequate because this problem is not about coverage (correctly predicted) by sets (as models) as much as it is about an example being covered by enough (possibly a majority) of sets. We adopt this particular measure and choice of ensemble style because of its ease of computation, especially in an incremental manner with respect to adding newer models

\begin{algorithm}
\KwData{$\hat{Y}$ (Prediction vectors from available models)}
$M \gets \phi$\\
 \While{$|\hat{Y}| > 0$}{
  $\hat{y}^* \gets \argmin_{\hat{y} \in \hat{Y}}{E(M \cup \hat{y})}$\\
  \eIf{ $E(M \cup \hat{y}^*) \le E(M) $}{
   $M \gets M \cup \hat{y}^*$\\
   $N \gets\ \hat{Y}-\hat{y}^*$\\
   }{
   break
  }
 }
 \KwResult{M}
 \caption{Greedy model subset selection: given a set of predictions from models, the algorithm selects a subset based on $E(M)$. We call this subroutine $E_{min}(M)$.}
\label{alg:greedy1}
\end{algorithm}

We propose an effective greedy algorithm to find the suitable subset of ensemble constituents in an efficient manner. 
Algorithm~\ref{alg:greedy1} iterates over all the predictions of all currently available models, and continues adding the predictions to the set of selected ones, $M$, as long as adding them improves $E(M)$. The advantage of this algorithm is its simplicity and efficiency, and it works well in practice. It is important to note that the simplicity of the formula chosen, enables the computation of $E(M+n)$ efficiently in an incremental manner, using $E(M)$, $n$ and certain bookkeeping by utilizing the definition of generalization error from Equation~\ref{eq:6}. We omit the details here. The algorithm subroutine is referred to as $E_{min}(M)$. 
This algorithm can be made more effective by the two following steps. First, apart from adding a new model to the bag once it satisfies $E(M \cup \hat{y}^*) \le E(M)$, we also consider dropping it, in spite of satisfying the condition. Second, (along with the first step) replacing $E(M \cup \hat{y}^*) \le E(M)$ with $E(M \cup \hat{y}^*) \le E(M) + \phi$, where $\phi$ is a small, positive constant. This enables growing the ensemble to a bigger size at the cost of a tiny increase in EGE in the short term. However, these make the $E_{min}$ algorithm more expensive. For the purpose of this paper, we will stick with the simpler Algorithm~\ref{alg:greedy1}. Note that this algorithm is somewhat similar but different in essential aspects compared to~\cite{ensembleselection}, which is based on a hill-climbing optimization after constructing an ensemble and computing validation score on a hold-out set, and is arguably slower for an incremental setting.

\section{Experiments}
%\subsection{Implementation Details}
%Our system has been implemented in Python, using Sklearn~\cite{scikit} for most of estimators, certain data preprocessing transformers. XGBoost, LightGBM are additional classifiers. Much of data processing and transformations happen through the Numpy library.

%We use the following set of 9 transformations as a default: {\em freq, pca, round, minmaxscaler, tanh, groupby+stddev, cbrt, sigmoid, stdscaler}. The default selection is based on historical performance. Additionally, other transformations can be readily specified, and more can be easily defined and added.
% {\em log, sqrt, square, cos, sin, tan, product, divide, sigmoid, abs, groupby+mean, groupby+min, groupby+max, extract\_day\_hour\_date, min, max, difference, numerical\_xor, cuberoot, cube}. 
%We use 5-fold cross validation for all our internal measurements and a default of {\em Random Forest, LightGBM} and {\em XGBoost} for both classification and regression.
Agent's exploration strategy was learned based on 62 datasets obtained from various open source repositories, primarily from OpenML~\footnote{\url{https://www.openml.org/}} and other propriety data. Different values of $t_{max}$ (5, 10, 15 ... 120 in minutes) were used during training; 10 transformations including feature selection, 4 estimators for classification and 3 in case of regression. We used learning rate parameter, $\alpha=0.05$, discount factor $\gamma=0.99$, and tried three values for $\epsilon$ (0.15,0.20,0.25), and finally used $\epsilon=0.2$ based on external validation. 

We evaluated our work on 56 binary classification data problems from OpenML. Note that these data are different from those used to train the agent.  We summarize all the results, while providing details for a subset of 10 amongst them. The subset was chosen based on the lowest 10 base Area under ROC curve for a plain random forest classifier on the data.% subset was made by  data to be set aside for evaluation was based on having at least 5,000 rows and at most 100,000, and at most 0.96 AUC value using default Random Forest Classifier, and then randomly selected 16 from the pool.
Our evaluation through out is based on splitting 33\% of the data for evaluation using a fixed seed (1) through Sklearn's train\_test\_split. The optimizers (ours or AutoSklearn) was only provided with the remaining 67\% of the rows and all numbers reported were based on testing on the unseen 33\%, using the metric Area under ROC curve. 
We report the relative reduction in AUC error compared to a base estimator as a measure of comparing across different models and aggregating results across different datasets. It is described as: $(1-AUC_{base}) - (1 - AUC_{opt}) \over 1-AUC_{base}$ . We additionally report a summary of results of a set of regression problems in terms of relative reduction of root mean squared error.

%
%\begin{table}[h]
%\centering
%\small
%\begin{tabular}{lll|lll}
%\hline
%Data             & Rows  & Cols & Data          & Rows  & Cols \\ \hline
%bank-marketing   & 45211 & 17   & elevators     & 16599 & 19   \\
%puma8NH          & 8192  & 9    & wind          & 6574  & 15   \\
%puma32H          & 8192  & 33   & electricity   & 45312 & 9    \\
%MagicTelescope   & 19020 & 11   & kin8nm        & 8192  & 9    \\
%delta\_elevators & 9517  & 7    & house\_8L     & 22784 & 9    \\
%numerai28.6      & 96320 & 22   & ailerons      & 13750 & 41   \\
%Satellite        & 5100  & 37   & bank32nh      & 8192  & 33   \\
%mc1              & 9466  & 39   & eeg-eye & 14980 & 15  \\
%\hline
%\end{tabular}
%\caption{OpenML Data used for evaluation.}
%\label{tab:data}
%\end{table}

\begin{table*}
\begin{tabular}{l|l|l|l|l|l|l|l|l|}
\cline{2-9}
                                           & \multicolumn{2}{l|}{\begin{tabular}[c]{@{}l@{}}Base Estimator \\ AUC\end{tabular}} & \multicolumn{2}{l|}{\begin{tabular}[c]{@{}l@{}}AUC after 30 \\ mins optimization\end{tabular}} & \multicolumn{2}{l|}{\begin{tabular}[c]{@{}l@{}}Reduction in error (1-AUC)\\ over Random Forest (\%age)\end{tabular}} & \multicolumn{2}{l|}{\begin{tabular}[c]{@{}l@{}}Reduction in error (1-AUC)\\ over XGBoost (\%age)\end{tabular}} \\ \hline 
\multicolumn{1}{|l|}{Classification data}                 & RF                                       & XGB                                     & ASKL                                           & APRL                                        & ASKL                                                 & APRL                                                        & ASKL                                              & APRL                                                     \\ \hline \hline
\multicolumn{1}{|l|}{pc2}                  & 0.5000                                   & 0.5000                                  & 0.8464                                         & \textbf{0.8823}                               & 69.29                                                & 76.46                                                         & 69.29                                             & 76.46                                                      \\ \hline
\multicolumn{1}{|l|}{numerai28.6}          & 0.5079                                   & 0.5174                                  & 0.5292                                         & \textbf{0.5605}                               & 4.33                                                 & 10.69                                                         & 2.44                                              & 8.92                                                       \\ \hline
\multicolumn{1}{|l|}{mc1}                  & 0.5789                                   & 0.5263                                  & 0.8211                                         & \textbf{0.8475}                               & 57.51                                                & 63.79                                                         & 62.23                                             & 67.81                                                      \\ \hline
\multicolumn{1}{|l|}{Hyperplane\_10\_1E-3} & 0.6618                                   & 0.6949                                  & 0.7344                                         & \textbf{0.7812}                               & 21.47                                                & 35.31                                                         & 12.94                                             & 28.29                                                      \\ \hline
\multicolumn{1}{|l|}{bank-marketing}       & 0.6634                                   & 0.6678                                  & 0.9272                                         & \textbf{0.9512}                               & 78.37                                                & 85.49                                                         & 78.08                                             & 85.30                                                      \\ \hline
\multicolumn{1}{|l|}{CreditCardSubset}     & 0.6665                                   & 0.6667                                  & \textbf{0.7383}                                & 0.7272                                        & 21.53                                                & 18.20                                                         & 21.50                                             & 18.17                                                      \\ \hline
\multicolumn{1}{|l|}{BNG(credit-g)}        & 0.7043                                   & 0.7242                                  & 0.8730                                         & \textbf{0.8952}                               & 57.06                                                & 64.57                                                         & 53.96                                             & 62.01                                                      \\ \hline
\multicolumn{1}{|l|}{bank32nh}             & 0.7061                                   & 0.7809                                  & 0.8980                                         & \textbf{0.9169}                               & 65.28                                                & 71.73                                                         & 53.43                                             & 62.08                                                      \\ \hline
\multicolumn{1}{|l|}{puma32H}              & 0.7898                                   & 0.8808                                  & 0.9617                                         & \textbf{0.9883}                               & 81.76                                                & 94.42                                                         & 67.84                                             & 90.15                                                      \\ \hline
\multicolumn{1}{|l|}{kin8nm}               & 0.8055                                   & 0.7829                                  & 0.9664                                         & \textbf{0.9741}                               & 82.72                                                & 86.68                                                         & 84.52                                             & 88.07                                                      \\ \hline \hline
\multicolumn{5}{|l|}{Mean error reduction (above ten datasets)}                                                                                                                                                                  & 53.93                                                & \textbf{60.73}                                                & 50.62                                             & \textbf{58.73}                                             \\ \hline
\multicolumn{5}{|l|}{Median error reduction (above ten datasets)}                                                                                                                                                                & 57.29                                                & \textbf{64.18}                                                & 53.70                                             & \textbf{62.05}                                             \\ \hline
\multicolumn{5}{|l|}{Mean error reduction (all 56 datasets)}                                                                                                                                                                     & 64.98                                                & \textbf{71.10}                                                & 62.07                                             & \textbf{68.4}                                              \\ \hline
\multicolumn{5}{|l|}{Median error reduction (all 56 datasets)}                                                                                                                                                                   & 68.32                                                & \textbf{73.54}                                                & 66.15                                             & \textbf{76.59}                                             \\ \hline
\end{tabular}
\caption{Comparing the accuracy of Random Forest, XGBoost Classifier with AutoSklearn and APRL ran for 30 minutes each. The fractional improvement is reported over reduction in error (1 - AUC) over the corresponding base model's.}
\label{tab:main}
\end{table*}

\subsection{Comparison to base estimators and Autosklearn}
Table~\ref{tab:main} presents the AUC scores for Random Forest and XGBoost algorithms upon running APRL and AutoSklearn for 30 minutes each, for 10 binary classification datasets. It compares APRL and AutoSklearn based upon reduction in the initial error (1-AUC) over each base estimator. It can be seen that APRL consistently achieves lower modeling error rates. Additionally, results for 56 binary (including the 10) classification datasets are summarized. 

\subsection{Optimization Performance with Varying Time} 
\begin{figure}[h]
    \centering
    \includegraphics[width=0.49\textwidth]{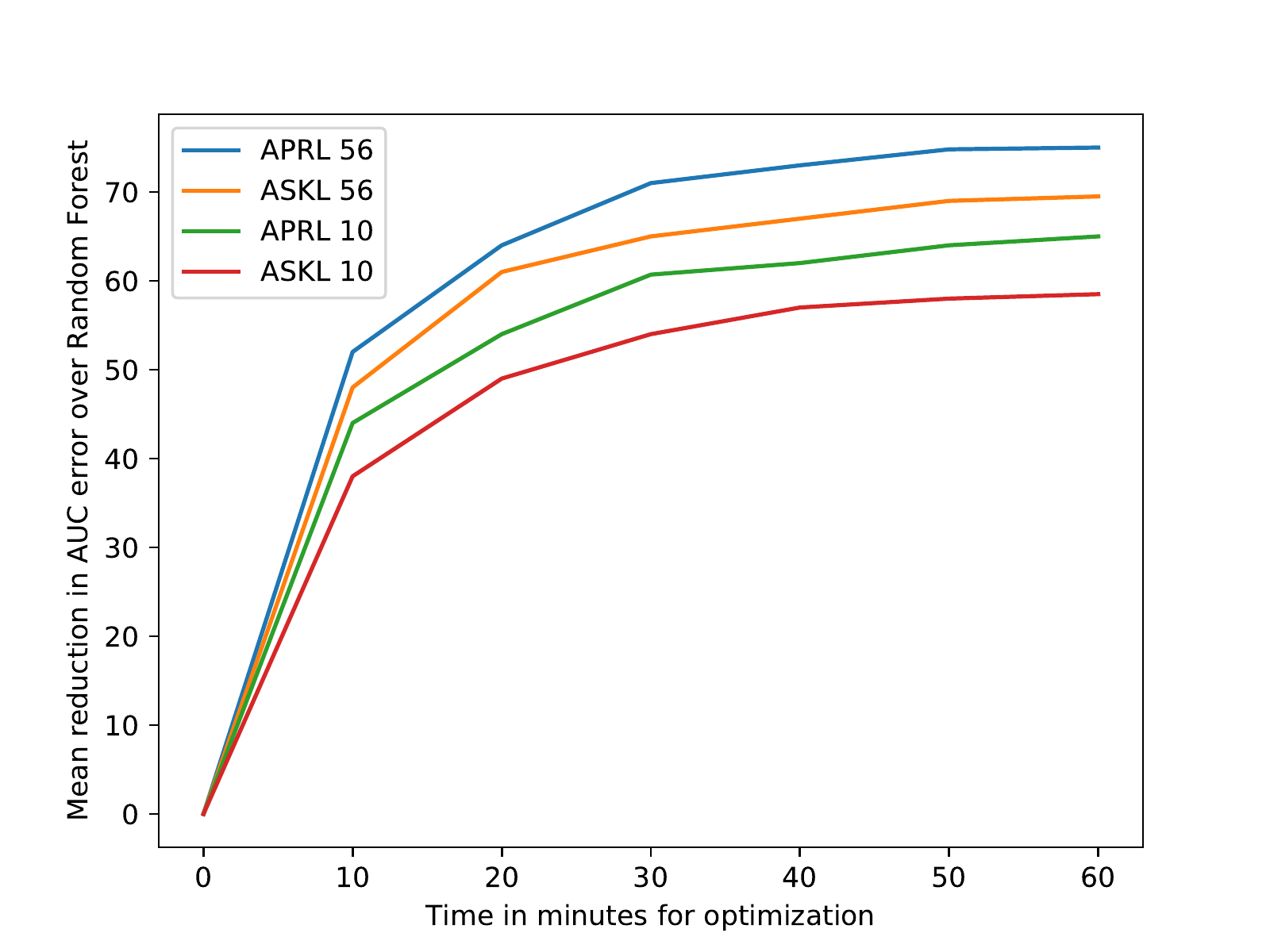}
    \caption{Mean reduction in AUC error (for 10 datasets in Table 1 and for all 56 datasets) due to AutoSklearn and APRL optimization for varying amounts of time (10 to 60 mins).}
    \label{fig:opt_time}
\end{figure}

Figure~\ref{fig:opt_time} displays the mean reduction in AUC error for the 56 and 10 set classification data for AutoSklearn and APRL over varying time constraints for optimization. In all cases, we can see increasing but diminishing returns with increasing time. Also, APRL performs better in each case.

\subsection{APRL vs FE alone vs FE+Ensemble}
For the 10 datasets listed in Table 1, upon allotting 30 minutes of runtime for each of the following: (a) APRL was able to reduce AUC error by 61\% compared to (b) 36\% of feature engineering alone (using XGBoost) on our implementation of the system described in~\cite{khuranaaaai18}, and (c) 44\% for ensemble using Algorithm 1 on feature engineering solution from part-b.
\subsection{Regression performance}
For 9 regression datasets, APRL was able to reduce root mean squared error by 31\% over Random Forest Regressor on an average, compared to 19\% reduction by AutoSklearn over Random Forest Regressor, by running each for 30 minutes. 
\section{Conclusion}
In this paper, we presented a novel automated machine learning framework called APRL, that produces effective ensembles using multiple data transformations, estimators and hyper-parameter optimization steps. At the heart of APRL is an autonomous agent that performs time-aware contextual decision making of various data science operations. It was constructed by learning an efficient strategy for data science exploration through reinforcement learning on historical datasets. To the best of our knowledge, it is the first of its kind where exploration is primarily guided by the need to optimize ensembles, as well as the first one to use reinforcement learning. We demonstrated the impact of our method by showing benefits against plain models, feature engineering alone, and ensembles constructed post feature engineering. In a comparison to Auto-Sklearn, it proved more effective in improving AUC for classification problems, under similar time constraints. %In the future, we plan to extend this framework to include more data exploration tools such as subsampling.
% Our future work involves integration of hyper-parameter optimization into this framework and to utilize more complex means of ensemble methods and error estimations.

%\clearpage
\bibliographystyle{plain}
\bibliography{ijcai19} 

\end{document}